\documentclass[runningheads]{llncs}
\usepackage[T1]{fontenc}

\usepackage{graphicx}

\usepackage{rotating}

\usepackage{makecell}
\usepackage{multirow}
\usepackage{pifont}
\usepackage{xcolor}

\newcommand{\cellwidth}{0.1\textwidth}
\newcommand{\cellwidthf}{0.15\textwidth}

\usepackage{array}

\begin{document}

\newcolumntype{P}[1]{>{\centering\arraybackslash}p{#1}}

\title{Leveraging Large Language Models for Topic Classification in the Domain of Public Affairs}
\titlerunning{Leveraging Large Language Models in the Domain of Public Affairs}
%
\author{Alejandro Peña\inst{1}\orcidID{0000-0001-6907-5826}, Aythami Morales\inst{1}\orcidID{0000-0002-7268-4785}, Julian Fierrez\inst{1}\orcidID{0000-0002-6343-5656}, Ignacio Serna\inst{1}\orcidID{0000-0003-3527-4071}, Javier Ortega-Garcia\inst{1}\orcidID{0000-0003-0557-1948},
Íñigo Puente\inst{2}, Jorge Córdova\inst{2}, Gonzalo Córdova\inst{2}}
\authorrunning{A. Peña, A. Morales, J. Fierrez, et al.}
\institute{BiDA - Lab, Universidad Autónoma de Madrid (UAM), Madrid 28049, Spain\\
\and VINCES Consulting, Madrid 28010, Spain}

%
\maketitle              
\begin{abstract}

The analysis of public affairs documents is crucial for citizens as it promotes transparency, accountability, and informed decision-making. It allows citizens to understand government policies, participate in public discourse, and hold representatives accountable. This is crucial, and sometimes a matter of life or death, for companies whose operation depend on certain regulations. Large Language Models (LLMs) have the potential to greatly enhance the analysis of public affairs documents by effectively processing and understanding the complex language used in such documents. In this work, we analyze the performance of LLMs in classifying public affairs documents. As a natural multi-label task, the classification of these documents presents important challenges. In this work, we use a regex-powered tool to collect a database of public affairs documents with more than $33$K samples and $22.5$M tokens. Our experiments assess the performance of $4$ different Spanish LLMs to classify up to $30$ different topics in the data in different configurations. The results shows that LLMs can be of great use to process domain-specific documents, such as those in the domain of public affairs.

\end{abstract}
\section{Introduction}


The introduction of the Transfomer model~\cite{vaswani2017attention} in early $2017$ supposed a revolution in the Natural Language Domain. In that work, Vaswani \textit{et al.} demonstrated that an Encoder-Decoder architecture combined with an Attention Mechanism can increase the performance of Language Models in several tasks, compared to recurrent models such as LSTM~\cite{hochreiter1997lstm}. Over the past few years, there has been a significant development of transformer-based language model architectures, which are commonly known as Large Language Models (LLM). Its deployment sparked a tremendous interest and exploration in numerous domains, including chatbots (e.g., ChatGPT,\footnote{https://openai.com/blog/chatgpt} Bard,\footnote{https://blog.google/technology/ai/bard-google-ai-search-updates/} or Claude\footnote{https://www.anthropic.com/index/introducing-claude}), content generation~\cite{gpt3,radford2019language}, virtual AI assistants (e.g., JARVIS~\cite{shen2023hugginggpt}, or GitHub's Copilot\footnote{https://github.com/features/preview/copilot-x}), and other language-based tasks~\cite{kenton2019bert}\cite{lewis2020bart}\cite{liu2019roberta}. These models address scalability challenges while providing significant language understanding and generation abilities. That deployment of large language models has propelled advancements in conversational AI, automated content creation, and improved language understanding across various applications, shaping a new landscape of NLP research and development. There are even voices raising the possibility that most recent foundational models~\cite{anil2023palm2}\cite{openai2023gpt4}\cite{instructgpt}\cite{touvron2023llama} may be a first step of an artificial general intelligence~\cite{bubeck2023sparks}.

Large language models have the potential to greatly enhance the analysis of public affairs documents. These models can effectively process and understand the complex language used in such documents. By leveraging their vast knowledge and contextual understanding, large language models can help to extract key information, identify relevant topics, and perform sentiment analysis within these documents. They can assist in summarizing lengthy texts, categorizing them into specific themes or subject areas, and identifying relationships and patterns between different documents. Additionally, these models can aid in identifying influential stakeholders, tracking changes in public sentiment over time, and detecting emerging trends or issues within the domain of public affairs. By leveraging the power of large language models, organizations and policymakers can gain valuable insights from public affairs documents, enabling informed decision-making, policy formulation, and effective communication strategies. The analysis of public affairs documents is also important for citizens as it promotes transparency, accountability, and informed decision-making. 

Public affairs documents often cover a wide range of topics, including policy issues, legislative updates, government initiatives, social programs, and public opinion. These documents can address various aspects of public administration, governance, and societal concerns. The automatic analysis of public affairs text can be considered a multi-label classification problem.  Multi-label classification enables the categorization of these documents into multiple relevant topics, allowing for a more nuanced understanding of their content. By employing multi-label classification techniques, such as text categorization algorithms, public affairs documents can be accurately labeled with multiple attributes, facilitating efficient information retrieval, analysis, and decision-making processes in the field of public affairs. 

\begin{figure}[htb]
    \centering
    \includegraphics[width=\textwidth]{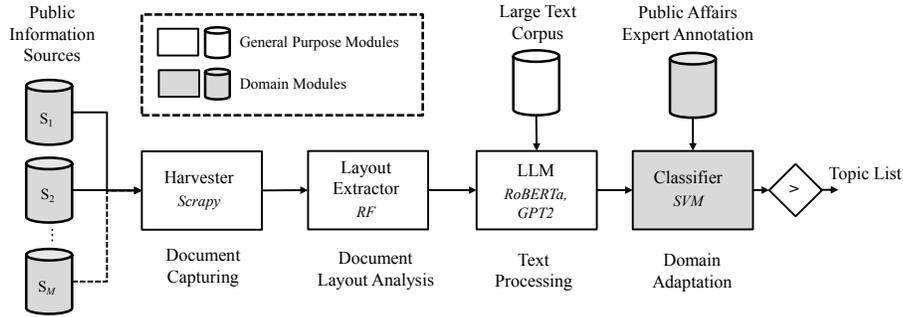}
    \caption{Block diagram of an automatic public affairs document analysis system. The white blocks represent general-purpose modules, while the grey blocks represent domain-specific modules.}
    \label{fig:project_diagram}
\end{figure}

This work focuses on NLP-related developments in an ongoing research project. The project aims to improve the automatic analysis of public affairs documents using recent advancements in Document Layout Analysis (DLA) and Language Technologies. The objective of the project is to develop new tools that allow citizens and businesses to quickly access regulatory changes that affect their present and future operations. With this objective in mind, a system is being developed to monitor the publication of new regulations by public organizations The block diagram of the system is depicted in Figure~\ref{fig:project_diagram}. The system is composed of three main modules: \textit{i)} Harvester module based on web scrappers; \textit{ii)} a Document Layout Analysis (DLA) module; and \textit{iii)} a Text Processing module. The Harvester monitors a set of pre-defined information sources, and automatically downloads new documents in them. Then, the DLA module conducts a layout extraction process, where text blocks are characterized and automatically classified, using Random Forest models, into different semantic categories. Finally, a Text Processing module process the text blocks using LLMs technology to perfom multi-label topic classification, finally aggregating individual text predictions to infer the main topics of the document. 

The full system proposed in Figure~\ref{fig:project_diagram} serves us to adapt LLMs to analyze documents in the domain of public affairs. This adaptation is based on the dataset used in our experiments, generated in collaboration with experts in public affairs regulation. They annotated over $92$K texts using a semi-supervised process that included a regex-based tool. The database comprises texts related to more than $385$ different public affairs topics defined by experts.

From all the analysis tool that can be envisioned in the general framework depicted in Figure~\ref{fig:project_diagram}, in the present paper we focus in topic classification, with the necessary details of the Harverster needed to explain our datasets and interpret our topic classification results. Other modules such as the Layout Extractor are left for description elsewhere. 

Specifically, the main contributions of this work are: 

\begin{itemize}

\item Within the general document analysis system for analyzing public affairs documents depicted in in Figure~\ref{fig:project_diagram}, we propose, develop, and evaluate a novel functionality for multi-label topic classification.

\item We present a new dataset of public affairs documents annotated by topic with more than 33K text samples and 22.5M tokens representing the main Spanish legislative activity between 2019 and 2022.

\item We provide experimental evidence of the proposed multi-label topic classification functionality over that new dataset using four different LLMs (including RoBERTa~\cite{liu2019roberta} and GPT$2$~\cite{radford2019language}) followed by multiple classifiers.

\end{itemize}

Our results shows that using a LLM backbone in combination with SVM classifiers suppose an useful strategy to conduct the multi-label topic classification task in the domain of public affairs with accuracies over $85\%$. The SVM classification improves accuracies consistently, even with classes that have a lower number of samples (e.g., less than $500$ samples).


The rest of the paper is structured as follows: In Section~\ref{sec:dataset} we describe  the data collected for this work, including data preprocessing details. Section~\ref{sec:methodology} describes the development of the proposed topic classification functionality. Section~\ref{sec:experiments} presents the experiments and results of this work. Finally, Section~\ref{sec:conclusions} summarizes the main conclusions.

\section{Data Collection and Analysis}
\label{sec:dataset}

The major decisions and events resulting from the legislative, judicial and administrative activity of public administrations are public data. Is a common practice, and even a legal requisite, for these administrations to publish this information in different formats, such as govermental websites or official gazettes\footnote{https://op.europa.eu/en/web/forum}. Here, we use a regex-powered tool to follow up parliamentary initiatives from the Spanish Parlament, resulting in a legislative-activities text corpora in Spanish. Parliamentary initiatives involve a diverse variety of parliament interactions, such as questions to the government members, legislative proposals, etc.

\begin{figure}[t!]
    \centering
    \includegraphics[width=0.9\textwidth]{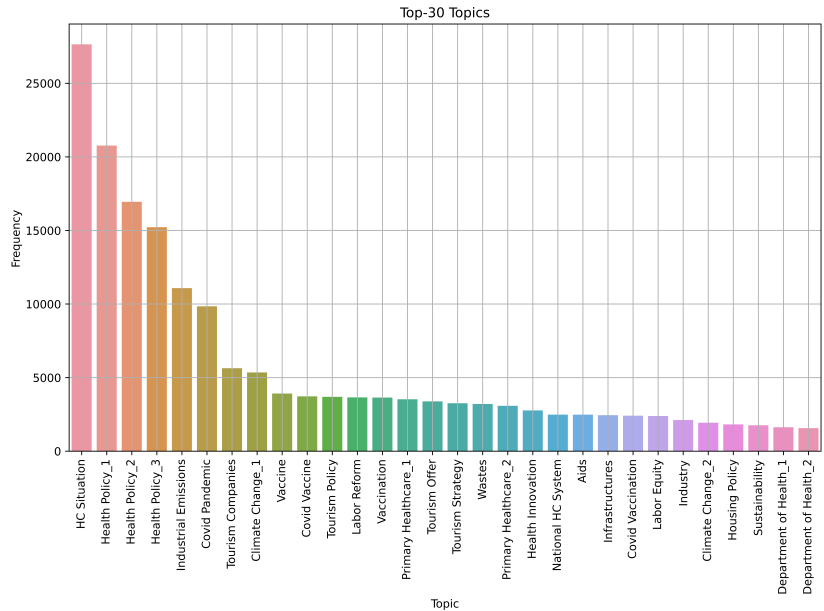}
    \caption{Distribution of the top 30 most frequent topics in the raw data.}
    \label{fig:top_concepts}
\end{figure}

Raw data were collected and processed with this tool, and comprise initiatives ranging from November $2019$ to October $2022$. The data is composed of short texts, which may be annotated with multiple labels. Each label includes, among others, topic annotations based on the content of the text. These annotations were generated using regex logic based on class-specific predefined keywords. Both topic classes and their corresponding keywords were defined by a group of experts in public affairs regulations. It is important to note that the same topic (e.g., ``\textit{Health Policy}'') can be categorized differently depending on the user's perspective (e.g., citizens, companies, governmental agencies). We have simplified the annotation, adding a ID number depending on the perspective used (e.g., ``\textit{Health Policy}$\mathit{\_1}$'' or ``\textit{Health Policy}$\mathit{\_2}$''). Our raw data is composed of $450$K initiatives grouped in $155$ weekly-duration sessions, with a total number of topic classes up to $385$. Of these $450$K samples, only $92.5$K were labeled, which suppose roughly $20.5\%$ of the samples. However, almost half of these are annotated with more than one label (i.e. $45.5$K, $10.06\%$ of samples), with a total number of labels of $240$K. Figure~\ref{fig:top_concepts} presents the distribution of the $30$ most frequent topics in the data, where we can clearly observe the significant imbalance between classes. The most frequent topic in the raw data is ``\textit{Healthcare Situation}'', appearing in more then $25$K data samples. Other topics, such as ``\textit{Health Policy}'', have an important presence in the data as well. However, only $8$ out of these $30$ topics reach $5$K samples, and only $5$ of them are present in at least $10$K. This imbalance, along with the bias towards health-related subjects in the most frequent topics, is inherent to the temporal framework of the database, as the Covid-$19$ pandemic situation has dominated significant public affairs over the past 3 years. Note that Figure~\ref{fig:top_concepts} depicts the thirty most frequent topics, whereas $385$ topics are present in the data. To prevent the effects of major class imbalances, we will now focus on the $30$ topics of Figure~\ref{fig:top_concepts}.

\subsection{Data Curation}
\label{sec:data_cleaning}

We applied a data cleaning process to the raw corpora to generate a clean version of the labeled data. We started by removing duplicated texts, along with data samples with less than $100$ characters. Some works addressing Spanish models applied a similar filtering strategy with a threshold of $200$ characters~\cite{raffel2020exploring,xue2021mt5,serrano2022rigoberta} with the aim of obtaining a clean corpus to pre-train transformer models. Here we set the threshold to $100$, as our problem here does not require us to be that strict (i.e., we do not want to train a transformer from scratch). Instead, we desired to remove extremely short text, which we qualitative assessed that were mainly half sentences, while retaining as much data as possible. In this sense, we filter text samples of any length starting with lowercase, to prevent half sentences to leak in. We also identified bad quality/noisy text samples to start with ``CSV'' or ``núm'', so we remove samples based on this rule. Finally, given the existence of co-official languages different from Spanish in Spain (e.g., Basque, Galician or Catalan), which are used by a significant percentage of Spanish citizens, we filter data samples from these languages. Due to the lack of reliable language detectors in these co-official languages, and the use of some linguistic, domain-specific patterns in the parliamentary initiatives, we identified a set of words in these languages and use it to detect and filter out potential samples not written in Spanish. We applied this process several times to refine the set of words.

At data sample level, we clean texts by removing excessive white spaces and initiative identifiers in the samples. We then filter URLs and non-alphanumeric characters, retaining commonly used punctuation characters in Spanish written text (i.e., ()-.¿?¡!$\_$;). After applying all the data curation process, we obtain a multi-label corpus of $33$,$147$ data samples, with annotations on the $30$ topics commented above. Table~\ref{tab:clean_corpus} presents the number of samples per topic category. Note that the number of samples of each topic has significantly decreased compared to the proportions observed in the raw data (see Figure~\ref{fig:top_concepts}). The impact of the data curation process is different between topics, leading to some changes in the frequency-based order of the topics. The topic with most data samples in the curated corpus is still ``\textit{Healthcare Situation}'', but the number of samples annotated with this topic has been reduced by half. On the other hand, we have several topics with less than $1$K samples, setting a lower limit of $518$.

\begin{table}[t!]
\scriptsize
\resizebox{\textwidth}{!}{
    \centering\setlength\doublerulesep{.1cm} 
    \begin{tabular}{ccc||ccc}
    \hline
    \textbf{ID} & \textbf{Topic}&\textbf{$\#$Samples} & \textbf{ID} & \textbf{Topic}& \textbf{$\#$Samples}\\
    \hline\hline
    $1$ & Healthcare Situation &$13561$ & $16$ & Primary Healthcare$\_1$&$1425$\\
    $2$ & Health Policy$\_1$ &$12029$ & $17$ & Sustainability&$1370$\\
    $3$ & Health Policy$\_2$&$8229$ & $18$ & Wastes&$1294$\\
    $4$ & Health Policy$\_3$&$8111$ & $19$ & Aids&$1216$\\
    $5$ & Industrial Emissions &$5101$ & $20$ & Primary Healthcare$\_2$&$1189$\\
    $6$ & Covid-$19$ Pandemic &$3298$ & $21$ & Tourism Offer&$1181$\\
    $7$ & Tourism Policy &$2209$ & $22$ &Labor Equity&$1074$\\
    $8$ & Tourism Companies &$2033$ & $23$ & Industry&$1051$\\
    $9$ & Climate Change$\_1$&$1930$ & $24$ & Infrastructures&$1029$\\
    $10$ & Vaccination &$1924$ & $25$ & Covid-$19$ Vaccination&$997$\\
    $11$ & Vaccine &$1751$ & $26$ & National Healthcare System&$964$\\
    $12$ & Covid-$19$ Vaccine&$1617$ & $27$ &Climate Change$\_2$ &$886$\\
    $13$ & Tourism Strategy&$1533$ & $28$ & Housing Policy&$744$\\
    $14$ & Labor Reform&$1529$ & $29$ &Department of Health$\_1$&$541$\\
    $15$ & Health Innovation&$1469$ & $30$ & Department of Health$\_2$&$518$\\
    \hline
    \end{tabular}}
    \caption{Summary of the parliamentary initiative database after the data cleaning process, which includes $33$,$147$ data samples with multi-label annotations across $30$ topics. We include a topic ID, the topic, and the number of samples annotated for each of them.}
    \label{tab:clean_corpus}
\end{table}

\section{Methodology and Models} 
\label{sec:methodology}

\begin{figure}[htb]
    \centering
    \includegraphics[width = 0.90\textwidth]{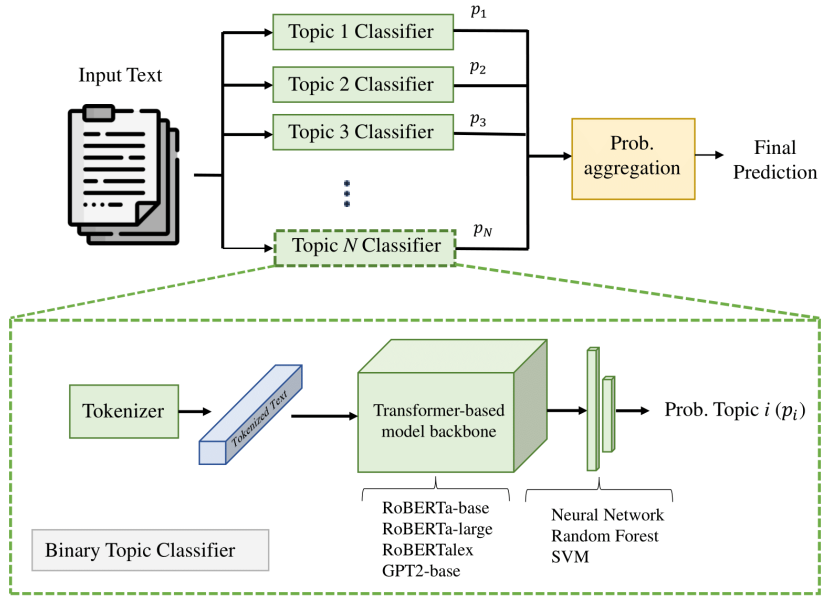}
    \caption{Proposed multi-label topic classification system, in which an individual topic detector is applied to an input text before aggregating all the predictions, and the architecture of each binary topic classifier.}
    \label{fig:system_architecture}
\end{figure}

As we previously mentioned in Section~\ref{sec:dataset}, the samples in our dataset may present more than one topic label. Hence, the topic classification task on this dataset is a multi-label classification problem, where we have a significant number of classes that are highly imbalanced. This scenario (i.e., high number of classes, some of them with few data samples, with overlapped subjects between classes) leads us to discard a single classifier for this task. Instead of addressing the problem as a multi-label task, we break it into small, binary detection tasks, where an individual topic detector is trained for each of the $30$ classes in a one vs all setup. This methodology, illustrated in Figure~\ref{fig:system_architecture}, represents a big advantage, as it provides us a high degree of versatility to select the best model configuration for each topic to deploy a real system. During inference, new data samples can be classified by aggregating the predictions of the individual classifiers~\cite{2018_INFFUS_MCSreview1_Fierrez}. 

The architecture of the binary topic models is depicted in Figure~\ref{fig:system_architecture}. We use a transformer-based model as backbone, followed by a Neural Network, Random Forest, or SVM classifier. In this work, we explore different transformer models, pretrained from scratch in Spanish by the Barcelona Supercomputing Center in the context of the MarIA project~\cite{gutierrez2021maria}. We included both encoder and decoder architectures. These model architectures are the following:

\begin{itemize}
    \item \textbf{RoBERTa-base}. An encoder-based model architecture with $12$ layers, $768$ hidden size, $12$ attention heads, and $125$M parameters.
    \item \textbf{RoBERTa-large}. An encoder-based model architecture with $24$ layers, $71$,$024$ hidden size, $16$ attention heads, and $334$M parameters.
    \item \textbf{RoBERTalex}. A version~\cite{gutierrezfandino2021legal} of RoBERTa-base, fine-tuned for the Spanish legal domain.
    \item \textbf{GPT$2$-base}. A decoder-based model architecture with $12$ layers, $768$ hidden size, $12$ attention heads, and $117$M parameters.
\end{itemize}

We listed above the configurations reported in~\cite{gutierrez2021maria} for the open-source models available in the the HuggingFace repository of the models.\footnote{https://huggingface.co/PlanTL-GOB-ES} The RoBERTa models~\cite{liu2019roberta} are versions of BERT models~\cite{kenton2019bert}, in which an optimized pre-training strategy and hyperparameter selection was applied, compared to the original BERT pre-training. The Spanish versions of these models were pre-trained following the original RoBERTa configuration, with a corpus of $570$ GB of clean Spanish written text. The RoBERTalex model is a fine-tuned version of Spanish RoBERTa-base, trained with a corpus of $8.9$ GB of legal text data. On the other hand, GPT$2$~\cite{radford2019language} is a decoder-based model of the GPT family~\cite{gpt3}\cite{openai2023gpt4}\cite{instructgpt}\cite{radford2018gpt}. As such, the model is aimed to generative tasks (note that modern versions of GPT models, such as InstructGPT~\cite{instructgpt} or GPT$4$~\cite{openai2023gpt4} are fine-tuned to follow human instructions, so they cannot be considered generative models in the same way as earlier GPT models), different from the RoBERTa family, which is specialized in text understanding. The version used of GPT$2$ was trained using the same corpus as the RoBERTa models. All the models use byte-level BPE tokenizer~\cite{radford2019language} with vocab size of $50$,$265$ tokens, and have the same length for the context windows, i.e. $512$. While left padding is used in the RoBERTa models, right padding is advisable for the GPT$2$ model.

\section{Experiments}
\label{sec:experiments}



As exposed in Section~\ref{sec:methodology}, due to the nature of the dataset collected for this work, we address multi-label topic classification by training a binary topic classifier for each class (one vs all), and then aggregating the individual predictions on a versatile way (e.g., providing rank statistics, topics over a fixed threshold, etc.). Hence, our experiments will focus on assessing the performance of different topic classifiers configurations, and the potential of the newly available Spanish language models in unconstrained scenarios (i.e., multi-label political data, with subjective annotations based on private-market interest). Section~\ref{sec:topic_classification} will evaluate first the performance of different transformer-based models on our dataset, and then explore the combination of the best-performance model with SVM and Random Forest classifiers.

We conduct all the experiments using a K-fold cross validation setup with $5$ folds, and report mean and average results between folds. We select True Positive Rate (TPR), and True Negative Rate (TNR) as our performance measures, due to the class imbalances in the parliamentary dataset. We use in our experiments the models available in the HuggingFace transformers library\footnote{https://huggingface.co/docs/transformers/index}, along with several sklearn tools. Regarding the hardware, we conducted the experiments in a PC with $2$ NVIDIA RTX $4090$ (with $24$ GB each), Intel Core i$9$, $32$GB RAM.

\subsection{Topic Classification in the Domain of Public Affairs}
\label{sec:topic_classification}

Recalling from Figure~\ref{fig:system_architecture}, our topic detector architecture is mainly composed of \textit{i)} a transformer backbone, and \textit{ii)} a classifier. We train the transformer models with a binary neural network classification output layer. For each topic, we train the detector using Weighted Cross Entropy Loss to address the class imbalance in a ``One vs All'' setup. Topic classifiers are trained for $5$ epochs using a batch size of $32$ samples, and freezing the transformer layers. Table~\ref{tab:backbone_comparison} presents the results of the topics classifiers using the four transformer models explored in this work (i.e., RoBERTa-base~\cite{gutierrez2021maria}, RoBERTa-large~\cite{gutierrez2021maria}, RoBERTalex~\cite{gutierrezfandino2021legal}, and GPT$2$-base~\cite{gutierrez2021maria}). We can observe a general behavior across the RoBERTa models. The classifiers trained for the topics with more samples obtain higher TPR means, close to the TNR mean values. In these cases, the classifiers are able to distinguish reasonably well text samples in which the trained topic is present. These results are, in general, consistent across folds, exhibiting moderate deviation values. This behavior degrades from Topic $9$ onwards, where the low number of samples (i.e., less than $2$K) leads to an increase of the TNR to values over $90\%$ with a decay of TPR. However, we can observe some exceptions in the classifiers using RoBERTa-base as backbone (topics $11$, $12$, $24$), where TNR scales to values close to $100\%$ while preserving TPR performances over $80\%$. Furthermore, RoBERTa-base classifiers exhibit better results than the RoBERTa-large classifiers (probably due to the constrained number of samples), and even than the RoBERTalex models. Remember that both RoBERTa-base and RoBERTalex are the same models, the latter being the RoBERTa-base model with a fine-tuning to the legal domain that, a priori, should make it more appropriate for the problem at hand. Regarding GPT$2$-based classifiers, we observe similar trends to those of the RoBERTa models, but exhibiting lower performances. This is not surprising, as the GPT model was trained for generative purposes, rather than text understanding like RoBERTa.

\begin{table}[t!]
\scriptsize
    \centering
    \begin{tabular}{P{\cellwidth}|P{\cellwidth}|P{\cellwidth}|P{\cellwidth}|P{\cellwidth}|P{\cellwidth}|P{\cellwidth}|P{\cellwidth}|P{\cellwidth}}
    \hline
    \multirow{2}{*}{\textbf{ID}}&\multicolumn{2}{c|}{\textbf{RoBERTa-b}~\cite{gutierrez2021maria}}&\multicolumn{2}{c|}{\textbf{RoBERTa-l}~\cite{gutierrez2021maria}}&\multicolumn{2}{c|}{\textbf{GPT$\mathbf{2}$-b}~\cite{gutierrez2021maria}}&\multicolumn{2}{c}{\textbf{RoBERTalex}~\cite{gutierrezfandino2021legal}}\\
    \cline{2-9}    &\textbf{TPR}&\textbf{TNR}&\textbf{TPR}&\textbf{TNR}&\textbf{TPR}&\textbf{TNR}&\textbf{TPR}&\textbf{TNR}\\
    \hline\hline
    $1$&$.80_{.07}$&$.75_{.19}$&$.78_{.08}$&$.76_{.19}$&$.58_{.14}$&$.60_{.15}$&$.79_{.10}$&$.70_{.18}$\\
    \hline
    $2$&$.87_{.09}$&$.88_{.04}$&$.84_{.11}$&$.86_{.05}$&$.61_{.25}$&$.82_{.05}$&$.83_{.10}$&$.82_{.06}$\\
    \hline
    $3$&$.83_{.08}$&$.87_{.04}$&$.81_{.09}$&$.87_{.04}$&$.65_{.18}$&$.79_{.07}$&$.79_{.09}$&$.84_{.05}$\\
    \hline
    $4$&$.86_{.07}$&$.89_{.03}$&$.83_{.10}$&$.88_{.03}$&$.69_{.17}$&$.79_{.07}$&$.80_{.10}$&$.86_{.04}$\\
    \hline
    $5$&$.76_{.05}$&$.81_{.06}$&$.72_{.07}$&$.81_{.07}$&$.63_{.06}$&$.74_{.09}$&$.67_{.08}$&$.80_{.06}$\\
    \hline
    $6$&$.82_{.05}$&$.87_{.02}$&$.83_{.05}$&$.87_{.03}$&$.67_{.04}$&$.63_{.08}$&$.68_{.06}$&$.83_{.04}$\\
    \hline
    $7$&$.85_{.04}$&$.93_{.03}$&$.83_{.06}$&$.91_{.05}$&$.64_{.08}$&$.78_{.08}$&$.75_{.07}$&$.94_{.03}$\\
    \hline
    $8$&$.82_{.02}$&$.89_{.05}$&$.81_{.03}$&$.88_{.06}$&$.63_{.04}$&$.78_{.07}$&$.69_{.02}$&$.91_{.04}$\\
    \hline
    $9$&$.79_{.10}$&$.90_{.04}$&$.77_{.11}$&$.89_{.06}$&$.58_{.08}$&$.76_{.08}$&$.68_{.07}$&$.91_{.03}$\\
    \hline
    $10$&$.76_{.26}$&$.96_{.03}$&$.67_{.31}$&$.95_{.03}$&$.49_{.42}$&$.91_{.10}$&$.62_{.34}$&$.95_{.04}$\\
    \hline
    $11$&$.89_{.11}$&$.98_{02}$&$.72_{.31}$&$.98_{.01}$&$.55_{.44}$&$.93_{.09}$&$.70_{.32}$&$.97_{.03}$\\
    \hline
    $12$&$.88_{.12}$&$.98_{.02}$&$.73_{.30}$&$.98_{.01}$&$.57_{.41}$&$.94_{.09}$&$.72_{.30}$&$.97_{.02}$\\
    \hline
    $13$&$.76_{.09}$&$.89_{.06}$&$.75_{.08}$&$.86_{.07}$&$.33_{.09}$&$.79_{.09}$&$.58_{.14}$&$.91_{.05}$\\
    \hline
    $14$&$.76_{.12}$&$.93_{.03}$&$.72_{.12}$&$.93_{.03}$&$.39_{.13}$&$.81_{.06}$&$.65_{.13}$&$.94_{.02}$\\
    \hline
    $15$&$.61_{.09}$&$.85_{.04}$&$.58_{.10}$&$.86_{.03}$&$.53_{.06}$&$.82_{.05}$&$.54_{.08}$&$.90_{.02}$\\
    \hline
    $16$&$.75_{.03}$&$.90_{.03}$&$.71_{.05}$&$.88_{.04}$&$.43_{.04}$&$.77_{.03}$&$.64_{.05}$&$.91_{.03}$\\
    \hline
    $17$&$.71_{.25}$&$.94_{.06}$&$.64_{.32}$&$.96_{.05}$&$.59_{.31}$&$.93_{.05}$&$.65_{.31}$&$.92_{.05}$\\
    \hline
    $18$&$.62_{.08}$&$.90_{.03}$&$.54_{.10}$&$.85_{.05}$&$.36_{.07}$&$.79_{.07}$&$.51_{05}$&$.91_{02}$\\
    \hline
    $19$&$.69_{.10}$&$.92_{.02}$&$.69_{.09}$&$.91_{.03}$&$.45_{.11}$&$.86_{.05}$&$.49_{.12}$&$.95_{.01}$\\
    \hline
    $20$&$.73_{05}$&$.93_{.02}$&$.73_{.06}$&$.90_{.03}$&$.32_{.04}$&$.86_{.03}$&$.58_{.05}$&$.94_{.02}$\\
    \hline
    $21$&$.67_{.04}$&$.89_{.05}$&$.67_{.06}$&$.86_{.06}$&$.48_{.05}$&$.84_{.05}$&$.45_{.03}$&$.93_{.03}$\\
    \hline
    $22$&$.71_{.05}$&$.95_{.02}$&$.66_{.03}$&$.94_{.02}$&$.40_{.04}$&$.89_{.04}$&$.51_{.03}$&$.97_{.01}$\\
    \hline
    $23$&$.70_{.08}$&$.96_{.02}$&$.57_{.17}$&$.96_{.02}$&$.24_{.17}$&$.96_{.01}$&$.43_{.17}$&$.98_{.01}$\\
    \hline
    $24$&$.83_{.08}$&$.97_{.04}$&$.69_{.11}$&$.98_{.01}$&$.20_{.24}$&$.98_{.01}$&$.55_{.18}$&$.98_{.01}$\\
    \hline
    $25$&$.80_{.16}$&$.97_{.04}$&$.54_{.36}$&$.97_{.04}$&$.44_{.40}$&$.98_{.03}$&$.57_{.34}$&$.97_{.04} $\\
    \hline
    $26$&$.52_{.10}$&$.95_{.01}$&$.48_{.13}$&$.96_{.01}$&$.17_{.03}$&$.96_{.02}$&$.40_{.10}$&$.98_{.01}$\\
    \hline
    $27$&$.72_{.08}$&$.97_{.02}$&$.62_{.08}$&$.97_{.02}$&$.25_{.07}$&$.97_{.01}$&$.56_{.05}$&$.98_{01}$\\
    \hline
    $28$&$.44_{.05}$&$.97_{.02}$&$.32_{.15}$&$.96_{.03}$&$0_{0}$&$1_{0}$&$.20_{.08}$&$.99_{01}$\\
    \hline
    $29$&$.46_{.06}$&$.98_{.01}$&$.17_{.04}$&$.99_{0}$&$0_{0}$&$1_{0}$&$.18_{04}$&$.99_{0}$\\
    \hline
    $30$&$.43_{.06}$&$.98_{.01}$&$.15_{.03}$&$.99_{0}$&$0_{0}$&$1_{0}$&$.15_{.03}$&$.99_{0}$\\
    \hline
    \end{tabular}
    \caption{Results of the binary classification for each topic (one vs all), using different transformer models with a Neural Network classifier. We report True Positive Rate (TPR) and True Negative Rate (TNR) as $\mathrm{mean}_\mathrm{std}$ (in parts per unit), computed after a K-fold cross validation ($5$ folds).}
    \label{tab:backbone_comparison}
\end{table}

It's worth noting here the case of Topic $1$, which obtains the lowest TNR mean value in all models, with deviation values over $0.15$, despite being the topic with more data samples (i.e. a third of the data). We hypothesize that the low performances when detecting negative samples is mostly due to the overlap with the rest of the topics, as this topic focuses on general healthcare-related aspects (remember from Table~\ref{tab:clean_corpus} that half of the topics are related with healthcare). 

\begin{table}[t!]
\scriptsize
    \centering
    \begin{tabular}{P{\cellwidthf}|P{\cellwidthf}|P{\cellwidthf}|P{\cellwidthf}|P{\cellwidthf}}
    \hline
    \multirow{2}{*}{\textbf{ID}}&\multicolumn{2}{c|}{\textbf{RoBERTa-b}~\cite{gutierrez2021maria} + \textbf{SVM}}&\multicolumn{2}{c|}{\textbf{RoBERTa-b}~\cite{gutierrez2021maria} + \textbf{RF}}\\
    \cline{2-5}    &\textbf{TPR}&\textbf{TNR}&\textbf{TPR}&\textbf{TNR}\\
    \hline\hline
    $1$&$.80_{.07}$&$.76_{.20}$&$.70_{.11}$&$.81_{.21}$\\
    \hline
    $2$&$.87_{.09}$&$.88_{.04}$&$.74_{.18}$&$.94_{.03}$\\
    \hline
    $3$&$.83_{.07}$&$.88_{.04}$&$.64_{.18}$&$.97_{.02}$\\
    \hline
    $4$&$.86_{.07}$&$.90_{.02}$&$.67_{.18}$&$.98_{.02}$\\
    \hline
    $5$&$.80_{.05}$&$.80_{.06}$&$.12_{.06}$&$.99_{.01}$\\
    \hline
    $6$&$.85_{.05}$&$.85_{.03}$&$.23_{.04}$&$.99_{0}$\\
    \hline
    $7$&$.90_{.02}$&$.90_{.05}$&$.49_{.06}$&$1_{0}$\\
    \hline
    $8$&$.89_{.01}$&$.86_{.07}$&$.33_{.02}$&$1_{0}$\\
    \hline
    $9$&$.88_{.07}$&$.87_{.04}$&$.24_{.05}$&$1_{0}$\\
    \hline
    $10$&$.84_{.18}$&$.94_{.03}$&$.51_{.41}$&$1_{0}$\\
    \hline
    $11$&$.92_{.08}$&$.97_{.02}$&$.57_{.38}$&$1_{0}$\\
    \hline
    $12$&$.93_{.07}$&$.98_{.02}$&$.56_{.36}$&$1_{0}$\\
    \hline
    $13$&$.87_{.04}$&$.85_{.08}$&$.08_{.02}$&$1_{0}$\\
    \hline
    $14$&$.87_{.07}$&$.88_{.04}$&$.14_{.04}$&$1_{0}$\\
    \hline
    $15$&$.70_{.08}$&$.80_{.06}$&$.06_{.02}$&$1_{0}$\\
    \hline
    $16$&$.89_{.03}$&$.88_{.04}$&$.13_{.05}$&$1_{0}$\\
    \hline
    $17$&$.79_{.18}$&$.92_{.06}$&$.59_{31}$&$1_{0}$\\
    \hline
    $18$&$.78_{.06}$&$.81_{.05}$&$.09_{.03}$&$1_{0}$\\
    \hline
    $19$&$.87_{.03}$&$.85_{.03}$&$.14_{.10}$&$1_{0}$\\
    \hline
    $20$&$.89_{.03}$&$.90_{.03}$&$.14_{.06}$&$1_{0}$\\
    \hline
    $21$&$.88_{.02}$&$.79_{.08}$&$.09_{.02}$&$1_{0}$\\
    \hline
    $22$&$.90_{.03}$&$.88_{.03}$&$.16_{.03}$&$1_{0}$\\
    \hline
    $23$&$.89_{.04}$&$.89_{.05}$&$.27_{.15}$&$1_{0}$\\
    \hline
    $24$&$.90_{.05}$&$.95_{.02}$&$.37_{.23}$&$1_{0}$\\
    \hline
    $25$&$.90_{.07}$&$.95_{.04}$&$.41_{.31}$&$.99_{.01}$\\
    \hline
    $26$&$.83_{.06}$&$.89_{.04}$&$.17_{.11}$&$1_{0}$\\
    \hline
    $27$&$.91_{.04}$&$.90_{.04}$&$.33_{.04}$&$1_{0}$\\
    \hline
    $28$&$.87_{.04}$&$.86_{.06}$&$.06_{.01}$&$1_{0}$\\
    \hline
    $29$&$.84_{.06}$&$.89_{.03}$&$.10_{.03}$&$1_{0}$\\
    \hline
    $30$&$.85_{.05}$&$.89_{.03}$&$.08_{.03}$&$1_{0}$\\
    \hline
    \end{tabular}
    \caption{Results of the binary classification for each topic (one vs all), using RoBERTa-base~\cite{gutierrez2021maria} in combination with SVM and Random Forest classifiers. We report True Positive Rate (TPR) and True Negative Rate (TNR) as $\mathrm{mean}_\mathrm{std}$ (in parts per unit), computed after a K-fold cross validation ($5$ folds).}
    \label{tab:svm_rf_results}
\end{table}

From the results presented in Table~\ref{tab:backbone_comparison}, we can conclude that RoBERTa-base is the best model backbone for our task. Now, we want to assess if a specialized classifier, such as Support Vector Machines (SVM) or Random Forests (RF), can be used to fine tune the performance to the specific domain. For these classifiers, we used RoBERTa-base as feature extractor to compute $768$-dimensional text embeddings from each of the text samples. We explored two approaches for these embeddings: \textit{i)} using the embedding computed for the [CLS] token, and \textit{ii)} averaging all the token embeddings (i.e., mean pooling). In the original BERT model~\cite{kenton2019bert}, and hence the RoBERTa model, the [CLS] is a special token appended at the start of the input, which the model uses during training for the Next Sentence Prediction objective. Thus, the output for this embedding is used for classification purposes, serving the [CLS] embedding as a text representation. We repeated the experiment using both types of representations, and end up selecting the first approach after exhibiting better results. Table~\ref{tab:svm_rf_results} presents the results of the topic models using RoBERTa-base text embeddings together with a SVM and Random Forest classifier. In all cases, we use a complexity parameter of $1$ and RBF kernel for the SVM, and a max depth of $1$,$000$ for the Random Forest. We note that these parameters can be tuned for each topic to improve the results. The first thing we notice in Table~\ref{tab:svm_rf_results} is the poor performance of the RF-based classifiers, which are the worst among all the configurations. Almost for all the topics under $2$K samples, the TNR saturates to $1$, and the TPR tends to extremely low values. From this, we can interpret that the classifier is not learning, and just predicting the negative, overrepresented class. However, the performance on the topics over $2$K samples is far from the one observed for the RoBERTa models of Table~\ref{tab:backbone_comparison}.  This could be expected, as the RF classifier is not the best approach to work with input data representing a structured vector subspace with semantic meaning, such as text/word embedding subspaces, specially when the number of data samples is low. On the other hand, the SVM performance clearly surpass all previous configurations in terms of TPR. While the results are comparable with those of RoBERTa-base 
 with NN for the first $5$ topics, this behavior is maintained for all topics, regardless of the number of data samples. Almost all classifiers achieve a TPR over $80\%$, except for topics $15$, $17$ and $18$.  Nevertheless, the results in these topics increase with the SVM (e.g., for topic $15$, where RoBERTa-base with the NN classifier achieved a TPR mean of $61\%$, here we obtain a $70\%$). TNR values are, in general, slightly lower, but this could be caused because in previous configurations, topic classifiers tend to exhibit bias towards the negative class as the number of samples falls (i.e., similar to the behavior of the RF classifier). Interestingly, the high deviation observed in the Topic $1$ TNR appears too in both SVM and RF classifiers, which could support our previous hypothesis. As we commented before, we suspect that an hyperparameter tuning could improve even more the SVM results on our data.

\section{Conclusions}
\label{sec:conclusions}

This work applies and evaluates Large Language Models (LLMs) for topic classification in public affairs documents. These documents are of special relevance for both citizens and companies, as they contain the basis of all legislative updates, social programs, public announcements, etc. Thus, enhancing the analysis of public documents using the recent advances of the NLP community is desirable.

To this aim, we collected a Spanish text corpora of public affairs documents, using a regex-powered tool to process and annotate legislative initiatives from the Spanish Parlament during a capture period over 2 years. The raw text corpora is composed of more than $450$K initiatives, with $92$K of them being annotated in a multi-label scenario with up to $385$ different topics. Topic classes were defined by experts in public affairs regulations. We preprocess this corpus and generate a clean version of more than $33$K multi-label texts, including annotations for the $30$ most frequent topics in the data.

We use this dataset to assess the performance of recent Spanish LLMs~\cite{gutierrezfandino2021legal}\cite{gutierrez2021maria} to perform multi-label topic classification in the domain of public affairs. Our experiments include text understanding models (three different RoBERTa-based models~\cite{liu2019roberta}) and generative models~\cite{radford2019language}, in combination with three different classifiers (i.e., Neural Networks, Random Forests, and SVMs). The results show how text understanding models with SVM classifiers supposes an effective strategy for the topic classification task in this domain, even in situations where the number of data samples is limited.

As future work, we plan to study in more depth biases and imbalances \cite{deAlcala2023n-sigma} like the ones mentioned before presenting Figure~\ref{fig:top_concepts}, and compensating them with imbalance-aware machine learning procedures \cite{Serna2022ai}. More recent LLMs can be also tested for this task, including multilingual and instruction-based models, which have shown great capacities in multiple NLP tasks, even in zero-shot scenarios. We will also continue our research by exploring the incorporation of other NLP tasks (e.g. text summarization, named entity recognition) and multimodal methods \cite{2023_SNCS_multiAI} to our framework, with the objective of enhancing automatic analysis of public affairs documents.

\section{Acknowledgments}

This work was supported by VINCES Consulting under the project VINCESAI-ARGOS and BBforTAI (PID$2021$-$127641$OB-I$00$ MICINN/FEDER). The work of A. Peña is supported by a FPU Fellowship (FPU$21$/$00535$) by the Spanish MIU. Also, I. Serna is supported by a FPI Fellowship from the UAM.

%
%
%
\bibliographystyle{splncs04}
\bibliography{bibliography}

\end{document}